\documentclass[sigconf]{acmart}

\usepackage[algo2e,ruled,lined,boxed,commentsnumbered, noend, linesnumbered, procnumbered]{algorithm2e}
\usepackage[binary-units]{siunitx}
\usepackage[abbreviations]{foreign}
\usepackage{graphicx}
\usepackage{textcomp}
\usepackage{xspace}
\usepackage{booktabs} 
\usepackage{multirow}
\usepackage{cleveref}
\usepackage{flushend}

\usepackage{lipsum}  
\usepackage{tablefootnote}
\usepackage{multirow}
\usepackage{soul}

\usepackage[inline]{enumitem}

\usepackage{tikz}
\tikzset{mygrid/.pic = {
		\draw[thin, dotted] (-3,-3) grid (3,3);
		\draw[->] (-3,0) -- (3,0);
		\draw[->] (0,-3) -- (0,3);}} %
\newcommand{\sys}{\textsc{SkipTrain}\xspace}
\newcommand{\tround}{training\xspace}
\newcommand{\sround}{synchronization\xspace}
\newcommand{\ML}{\ac{ML}\xspace}
\newcommand{\FL}{\ac{FL}\xspace}
\newcommand{\DL}{\ac{DL}\xspace}

\newcommand{\femnist}{FEMNIST\xspace}

\newcommand{\cifar}{CIFAR-10\xspace}

\newcommand{\sgd}{{\xspace}\ac{SGD}\xspace}
\newcommand{\dpsgd}{{\xspace}\ac{D-PSGD}\xspace}
\newcommand{\iid}{\ac{IID}\xspace}
\newcommand{\niid}{\ac{non-IID}\xspace}

\newcommand{\xsym}{\boldsymbol{x}\xspace}
\newcommand{\constrained}{\textsc{constrained}\xspace}

\newcommand{\greedy}{\textsc{Greedy}\xspace} \graphicspath{ {figures/} }
\usepackage{tikz}
\usepackage{pgfplots}
\usepackage{pgfplotstable}
\usepackage[eulergreek]{sansmath}

\pgfplotsset{compat=newest}
\usepgfplotslibrary{external,units,colorbrewer,groupplots,fillbetween}
\tikzsetexternalprefix{figures/}
\tikzset{external/mode=list and make}
\usetikzlibrary{patterns}

\makeatletter
\begingroup\endlinechar=-1\relax
\everyeof{\noexpand}%
\edef\x{\endgroup\def\noexpand\homepath{%
		\@@input|"kpsewhich --var-value=HOME" }}\x
\makeatother

\def\overleafhome{/tmp}
\newcommand{\inputplot}[2]{%
	\ifx\homepath\overleafhome%
	\IfBeginWith{#1}{plots}{\includegraphics{main-figure#2.pdf}}{#1}%
	\else%
	{\sffamily\scriptsize\input{#1}}
	\fi
}

\newcommand{\newgroupwidth}[2]%
{\expandafter\xdef\csname groupwidth#1\endcsname{#2}}

\usepackage{acronym}
\acrodef{DL}{decentralized learning}
\acrodef{ML}{machine learning}
\acrodef{D-PSGD}{decentralized parallel stochastic gradient descent}
\acrodef{FL}{federated learning}
\acrodef{SGD}{stochastic gradient descent}
\acrodef{IID}{independent and identically distributed}
\acrodef{non-IID}{non independent and identically distributed}
\acrodef{RMSE}{root mean square error}
\acrodef{RMW}{random model walk}
\acrodef{GL}{gossip learning}
\acrodef{DWT}{discrete wavelet transform}
\acrodef{FFT}{fast Fourier transform}
\acrodef{IoT}{Internet-of-Things}
\acrodef{UAV}{unmanned aerial vehicles}
\acrodef{CNN}{Convolutional Neural Network} %

\renewcommand\footnotetextcopyrightpermission[1]{} 
\begin{document}

\title{Energy-Aware Decentralized Learning with Intermittent Model Training}

\author{Akash Dhasade$^{1}$, Paolo Dini$^{2}$, Elia Guerra$^{3}$, Anne-Marie Kermarrec$^1$, Marco Miozzo$^{2}$, Rafael Pires$^{1}$, Rishi Sharma$^{1}$, Martijn de Vos$^{1}$}
\def \authors{author one, author two, author three, author four}
\affiliation{%
\institution{$^1$EPFL
\country{Switzerland}}
}
\affiliation{%
  \institution{$^2$Centre Tecnològic de Telecomunicacions de Catalunya
  \country{Spain}}
}
\affiliation{%
\institution{$^3$Independent Researcher
\country{Italy}}
}
\renewcommand{\shortauthors}{Dhasade et al.}
\thanks{Work done while Elia Guerra was affiliated with CTTC during a research visit at EPFL.

This publication has been partially funded by European Union Horizon 2020 research and innovation programme under Grant Agreement No. 953775 (GREENEDGE)}
\begin{abstract}
\Ac{DL} offers a powerful framework where nodes collaboratively train models without sharing raw data and without the coordination of a central server. 
In the iterative rounds of \Ac{DL}, models are trained locally, shared with neighbors in the topology, and aggregated with other models received from neighbors.
Sharing and merging models contribute to convergence towards a consensus model that generalizes better across the collective data captured at training time.
In addition, the energy consumption while sharing and merging model parameters is negligible compared to the energy spent during the training phase.
Leveraging this fact, we present \sys, a novel \Ac{DL} algorithm, which minimizes energy consumption in decentralized learning by strategically \emph{skipping} some training rounds and substituting them with \emph{synchronization} rounds.
These training-silent periods, besides saving energy, also allow models to better mix and finally produce models with superior accuracy than typical \ac{DL} algorithms that train at every round.
Our empirical evaluations with 256 nodes demonstrate that \sys reduces energy consumption by 50\% and increases model accuracy by up to 12\% compared to D-PSGD, the conventional \Ac{DL} algorithm.
\end{abstract}

\keywords{decentralized learning, machine learning, energy efficiency, peer-to-peer}

\maketitle
\pagestyle{plain} 

\acresetall
\section{Introduction}
\label{sec:intro}

\Ac{DL} represents an attractive alternative to centralized \ac{ML}, as it addresses privacy concerns by not moving training data while eliminating the dependency on a central server~\cite{lian2017can,koloskova2020unified,dhasade2022tee}.
In each round, nodes independently train an \ac{ML} model on their private datasets and share their model updates with immediate neighbors for aggregation, according to the underlying communication topology. 
This iterative process of training, sharing, and aggregation continues until the model has reached convergence.
Notable advantages of \ac{DL} over traditional centralized training methods include the removal of traffic and computational bottlenecks on the centralized server, along with the communication saved for not having to move the training data~\cite{lian2017can,lian2018asynchronous}.
\ac{DL} has been employed in various application domains such as  healthcare~\cite{lu2020decentralized,kasyap2021privacy,tedeschini2022decentralized} and \ac{IoT}~\cite{gerz2022comparative,lian2022decentralized}.

One of the most celebrated algorithms for collaborative training in \DL is the \dpsgd algorithm~\cite{lian2017can}, where nodes iteratively perform training, sharing, and aggregation operations in synchronous rounds of communication.
Since the introduction of \dpsgd, several other \DL algorithms were proposed to address problems related to data heterogeneity (\ie, individual nodes with different data distributions)~\cite{hsieh2020non}, system heterogeneity (\ie, individual nodes with different hardware capabilities)~\cite{lian2018asynchronous} and communication overheads~\cite{strom2015scalable,alistarh2018convergence, shokri2015privacy, dhasade2023get}.

Another important, yet overlooked problem of \DL is the training energy consumed by \dpsgd and variants.
While energy consumption is an overarching issue in the field of \ac{ML}~\cite{garcia2019estimation}, the energy consumption of \ac{DL} algorithms is even more concerning since they typically take many more rounds to converge than centralized approaches~\cite{devos2023epidemic}.
Energy awareness is crucial in settings where nodes are battery-limited and can only participate in the learning process for so long, \eg, in \ac{IoT} environments~\cite{huang2022toward} or in \ac{UAV} swarms~\cite{qu2021decentralized}.

The different operations in the \ac{DL} life cycle consume different amounts of energy.
Typically, nodes in \ac{DL} algorithms perform the following operations each round: 
\begin{enumerate*}[label=\emph{(\roman*)}]
\item \label{l:train}locally training the model; 
\item \label{l:share}exchanging the model with neighbors; and
\item \label{l:merge}aggregating models received from neighbors.
\end{enumerate*}
Most of the energy consumption happens at training time~\ref{l:train}, while that of communication, \ie \ref{l:share} and \ref{l:merge}, remains low.
Specifically, using the model adopted by Guerra et al.~\cite{guerra2023cost} on a network of $256$ nodes performing \dpsgd on the \cifar dataset (\Cref{subsec:experimental_setup}), the training step consumes \SI{1.51}{\kilo\watt\hour}, while communication and aggregation accounts for just \SI{7}{\watt\hour}, \ie, training is more than $200\times$ costlier in terms of energy than sharing and aggregation.

\begin{figure}[tb!]
	\centering
	\includegraphics{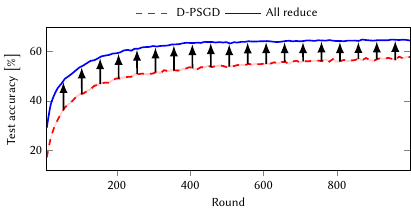}
	\caption{Comparison between \dpsgd (mean accuracy across nodes) and \dpsgd with all reduce (accuracy of the global average of models) on 256 nodes in a 6-regular topology. All-reduce significantly boosts model performance.}
    
 \label{fig:dpsgd_vs_dpsgd_all_reduce}
\end{figure}

From an energy perspective, increasing the amount of sharing and aggregation operations has a negligible impact on energy consumption.
Sharing and aggregating models combines the distinct contributions of individual nodes in \ac{DL}.
Ultimately, this leads the models towards a unique global consensus model, like the one produced by the central server in \ac{FL} or through a decentralized all-reduce operation~\cite{yu2022gadget}. 
Establishing an all-reduce synchronization across decentralized nodes enables faster propagation of local knowledge, which otherwise would require several hops between two distant nodes on the communication topology.
While achieving such synchronization in large-scale decentralized networks is impractical due to the massive communication volume required to perform all-to-all model exchange, it does significantly improve accuracy.
To quantify the benefits of additional synchronization, we show in \Cref{fig:dpsgd_vs_dpsgd_all_reduce} the results of standard \dpsgd with those hypothetically obtained by performing an all-reduce at every round. 
Evaluating the all-reduced model leads to an approximate improvement of 
\SI[scientific-notation=false]{10}{\%} over \dpsgd.
Yet, the situation is not strictly binary. 

While \dpsgd consistently trains all the time, \ie, associating training and sharing, intermediary alternatives between \dpsgd and the all-reduce strategy are possible.
This insight is the enabling element for our novel energy-aware \ac{DL} algorithm, named \sys, that significantly reduces the energy required to train an \ac{ML} model in \ac{DL} settings.
\sys is based on the observations that
\begin{enumerate*}[label=\emph{(\roman*)}]
	\item training is much more energy-hungry than sharing; and %
	\item sharing and aggregation result in more performant models. %
\end{enumerate*}
In \sys, nodes skip some training rounds (\ie, training, sharing, and aggregation) in favor of synchronization rounds (\ie, sharing and aggregation).
This significantly reduces the energy impact for the learning task while positively influencing model performance.

\sys comes in two variants, namely \sys and \sys-\constrained, to accommodate various scenarios and address different use cases.  
In \sys, nodes trade specific training rounds for \emph{sharing-only} ones. This not only significantly limits the energy consumption but also accelerates the model convergence.
\sys-\constrained extends \sys to operate in scenarios where nodes are subject to energy constraints, typically characterized by individual energy budgets, as commonly seen in \ac{IoT} devices or \acp{UAV}.
The main idea behind \sys-\constrained is that each node, depending on its energy capacity, makes an individual probabilistic decision in every round to either engage in  training or skip training to solely perform sharing. %
From the perspective of an individual node, this effectively redistributes its training budget across the whole learning duration.

We extensively evaluate the efficiency and performance of \sys on \niid data distributions using the \cifar and \femnist datasets, with several communication graphs consisting of $256$ nodes.
A unique aspect of our experimental setup is the integration of energy traces that we compiled by extending existing data~\cite{lai2022fedscale}. %
These energy traces serve as a crucial foundation for our evaluation, providing a realistic measure of the energy savings achieved by \sys in real-world scenarios and with hardware profiles of real devices.
Our empirical evaluations demonstrate that \sys achieves a $50\%$ reduction in energy consumption and increases model accuracy by up to $7$\% compared to D-PSGD.
In energy-constrained settings, \sys-\constrained increases model accuracy by up to $12$\% compared to \dpsgd.

In summary, we make the following contributions:
\begin{enumerate}
    \item We introduce \sys: an energy-aware learning approach that replaces training rounds with synchronization rounds to both reduce the energy spent during the training process and to boost model accuracy.
    
    \item We extend \sys by introducing \sys-\constrained that can be deployed in settings where each node has an energy capacity.

    \item We create traces containing the energy impact of different types of mobile phones, allowing us to evaluate the effectiveness of \sys under realistic, real-world conditions.

    \item We implement \sys and \sys-\constrained, and empirically compare its efficiency and performance in terms of energy consumption and achieved model accuracy with \dpsgd, using two standard image classification tasks, different communication graphs with 256 nodes, \niid data distributions, and realistic energy consumption traces. %
    \footnote{\url{https://github.com/sacs-epfl/SkipTrain.git}}

\end{enumerate}

We present next (\Cref{sec:background}) some background on decentralized learning, \dpsgd and the energy model we adopt.
\Cref{sec:skiptrain} describes \sys and \sys-\constrained in detail.
\Cref{sec:eval} evaluates the proposed algorithms on \cifar and \femnist datasets.
\Cref{sec:discussion,sec:related} discuss about the limitations of \sys and related work, before we conclude in \Cref{sec:conclusion}.

\section{Background}
\label{sec:background}
We first outline the \DL setup, then explain the \dpsgd algorithm, and finally describe how we can measure the energy consumption of individual nodes, which is required to generate our energy traces. %

\subsection{Decentralized learning}
\label{subsec:background_DL}
We consider a setting with $n$ nodes that seek to collaboratively learn an \ML model.
Each node has access to its local dataset drawn from its own local distribution $D_i$, potentially different across nodes. 
The goal of the training process is to find the parameters of the model $\xsym$ that perform well on the union of the local distributions $D=\cup_{i=1}^n D_i$ by minimizing the average loss function:
\begin{equation}
    \label{eqn:DL_obj}
    \min_{\xsym} f(\xsym) = \frac{1}{n}\sum_{i = 1}^n \underbrace{\mathbb{E}_{\xi \sim D_i}  [F_i(\xsym, \xi)]}_{:=f_i(\xsym)},
\end{equation}
where $f_i$ is the local objective function of node $i$ and $F_i(\xsym, \xi)$ is the loss of the model $\xsym$ on the sample $\xi \in D_i$. 
To collaboratively solve \Cref{eqn:DL_obj}, each node can exchange messages with its neighbors through a communication topology represented as an undirected graph $G=(V,E)$, where $V$ denotes the set of all nodes and $(i,j) \in E$ denotes an edge or communication channel between nodes $i$ and $j$. 

\subsection{\dpsgd algorithm}

The standard algorithm to solve the \DL task is \dpsgd~\cite{lian2017can}, illustrated in \Cref{alg:dpsgd}.
In a generic round $t<T$, each node $i$ executes $E$ local steps of \sgd by sampling mini-batches from local distribution $D_i$ (Lines~\ref{line:local_update_start}--\ref{line:local_update_end}).
Subsequently, it shares the updated model $\xsym_i^{t-\frac{1}{2}}$ with its neighbors (Line~\ref{line:share}). 
Finally, the node performs a weighted average of its local model with those of its neighbors through the matrix $W$ (Line~\ref{line:aggregate}) that encodes the strength of the connection between the nodes.
To ensure convergence towards a stationary point of \Cref{eqn:DL_obj}, the matrix $W\in \mathbb{R}^{n \times n}$ should be symmetric ($W_{i,j} = W_{j,i}$) and doubly stochastic (\ie, $\sum_{j \in V} W_{i,j} = 1$ and $\sum_{i \in V} W_{i,j} = 1$)~\cite{lian2017can}.
We obtain $W$ by computing Metropolis-Hastings weights~\cite{xiao2004fast} based on the network topology $G = (V,E)$ as follows:
\[
W_{i,j} = \left\{
    \begin{array}{ll}
     \frac{1}{\text{max}(\text{degree}(i), \text{degree}(j))+1}  & \text{if } i \neq j \text{ and } (i,j) \in E, \\
      1 - \sum_{i \neq j} W_{i,j}   & \text{if } i = j, \\
      0 & \text{otherwise.}
    \end{array}
    \right\}
\]

\begin{algorithm2e}
    \DontPrintSemicolon
    \caption{D-PSGD, Node $i$}
    \label{alg:dpsgd}
    Initialize $\xsym_i^0$\;
    \For{$t = 1, \dots, T$}{
        $\xsym_i^{t-\frac{1}{2}} \leftarrow \xsym_i^{t-1}$\;
        \For{$e = 1, \dots, E$}{\label{line:local_update_start}
            $\xi_i \leftarrow$ mini-batch of samples from $D_i$\;
            $\xsym_i^{t-\frac{1}{2}} \leftarrow \xsym_i^{t-\frac{1}{2}} - \eta \nabla f_i(\xsym_i^{t-\frac{1}{2}}, \xi_i)$\;
        }\label{line:local_update_end}
        Send $\xsym_i^{t-\frac{1}{2}}$ to the neighbors \label{line:share}\;
        $\xsym_i^t \leftarrow \sum_{j\in V} W_{j,i} \xsym_j^{t-\frac{1}{2}}$ \label{line:aggregate}\;
    }
    \KwRet $\xsym_i^T$\;
\end{algorithm2e}

\subsection{Energy model}
In this work, we focus on the energy consumption of the training process. Following the same approach of Guerra \etal~\cite{guerra2023cost}, the energy consumption of the training process for a node $i$, during a generic iteration $t$, is the product of the power consumption of the hardware, $P_{hw,i}^t$, and the duration of the task $\Delta_i^t$:
\begin{equation}
    \mathcal{E}_i^t = P_{hw, i}^t \Delta_i^t.
    \label{eq:energy_training_round}
\end{equation}
Therefore, the total energy consumption of all the nodes during $T$ rounds is given by:
\begin{equation}
    \mathcal{E} =  \sum_{t=1}^{T}\sum_{i \in V} \mathcal{E}_i^t.
    \label{eq:total_training_energy}
\end{equation}
Several tools are available to estimate the energy consumption of the training process, \eg, the Machine Learning emission calculator~\cite{lacoste2019quantifying} and Green algorithms~\cite{lannelongue2021green}.
These tools report values based on theoretical models without measuring the hardware power consumption in real time.
However, they are only compatible with particular hardware configurations.
In this work, we create energy traces for particular devices derived from a combination of the Burnout benchmark~\cite{ignatov:2022:burnout}, the AI benchmark~\cite{ignatov2018ai} and FedScale~\cite{lai2022fedscale}.
We discuss the construction of these traces in more detail in \Cref{subsec:experimental_setup}.

From an energy perspective, we consider both an unconstrained and a constrained setting.
In the first setting, we simply measure training energy following \Cref{eq:total_training_energy} and the energy traces of real devices.
In the energy-constrained setting, each node $i$ has a computational budget $\tau_i$ that defines the maximum number of training rounds that can be executed before its battery is depleted.

\section{\sys}
\label{sec:skiptrain}
We first introduce in~\Cref{subsec:sync_rounds} the \sys algorithm, which replaces energy-intensive training rounds with \emph{\sround rounds} to minimize energy usage and boost model accuracy.
Next, we explain the \sys-\constrained algorithm that targets an energy-constrained setting by introducing \emph{training probabilities}.

\begin{figure*}[t]
	\centering
    \includegraphics[width=\linewidth]{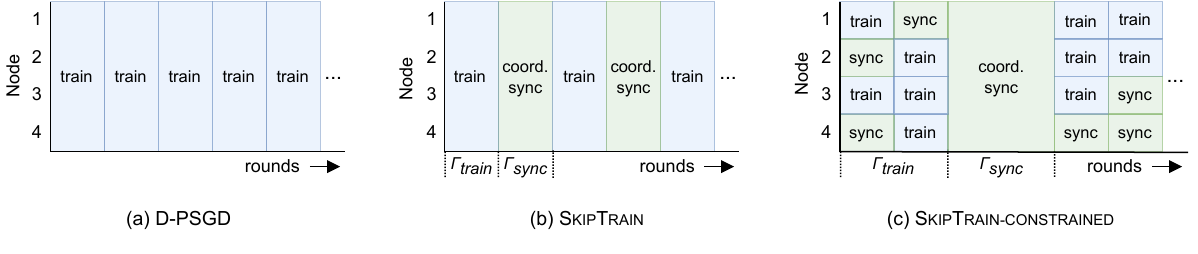}
    \caption{The operations performed by  of \dpsgd, \sys and \sys-\constrained during multiple rounds, for four nodes.}
	\label{fig:exp_comparison_fl}
\end{figure*}

\subsection{Coordinated \sround rounds}
\label{subsec:sync_rounds}
\dpsgd has a one-training-one-sharing approach, \ie nodes iteratively perform consecutive training and sharing steps. %
While \dpsgd eventually converges, it experiences weaker synchronization and higher variance between nodes, leading to slower convergence.  %
Effectively, \FL~\cite{mcmahan2017communication} and \dpsgd with fully-connected topologies show that strongly synchronized models lead to boosted convergence speeds~\cite{shi2022improving,li2022destress,scaman2017optimal}.
While fully-connected topologies have a high communication overhead,  approximate synchronization can be achieved by performing the synchronization step (sharing and aggregation) multiple times~\cite{xiao2004fast}.
Leveraging the fact that \sround rounds have a negligible impact from the perspective of energy required, we optimize the energy consumption of \dpsgd by replacing several \tround rounds with  \sround rounds, where the models are only shared and aggregated. 

In \sys, a \tround round is similar to a complete round in \dpsgd (\Cref{fig:exp_comparison_fl}a).
A node $i$ carries out the training on the local data, sharing of the model with neighbors, and aggregation of the received models.
During a \sround round, however, only the sharing and aggregation steps are executed.
\sys follows a pattern of alternating between a batch of $\Gamma_{\rm train}$ \tround rounds and $\Gamma_{\rm sync}$ \sround rounds (\Cref{fig:exp_comparison_fl}b).
The target is to alternate training and synchronization rounds such that the overall number of rounds does not increase when compared to standard \dpsgd.
As we show empirically in \Cref{sec:eval}, this leads to a better target accuracy for the same number of training rounds.
We detail in \Cref{subsec:hyperparameters_optimization} how we tune the values of $\Gamma_{\rm train}$ and $\Gamma_{\rm sync}$.

\subsection{Partial client participation}
\label{subsec:partial_client_participation}

In settings where devices run on batteries, nodes cannot perform arbitrarily many training rounds because of energy constraints.
Specifically, a node $i\in V$ can train for at most $\tau_i$ \tround rounds.
One way to incorporate these constraints in \dpsgd would be to carry out consecutive \tround rounds until the allocated energy budget is exhausted, followed by \sround-only rounds.
In other words, node $i$ with training budget $\tau_i$ will exclusively execute training rounds up to iteration $t = \tau_i$.
For the remaining rounds, the node will execute only synchronization rounds.  
We refer to this approach as \greedy and use this as a baseline.

While \sys with injected \sround rounds between \tround rounds is energy-efficient, energy budgets may still limit the number of \tround rounds that a node can perform.
Therefore, we extend \sys with \emph{training probabilities} and propose \sys-\constrained, specially crafted for energy-constrained settings.
In \sys-\constrained, nodes perform coordinated \sround rounds just like \sys.
However, in a \tround round, each node independently performs or skips training based on training probabilities derived from its own energy budgets~(\Cref{fig:exp_comparison_fl}c). 

Specifically, if $T$ is the total number of rounds executed by \sys, the maximum number of training rounds that a node executes is given by:
\begin{equation}
    T_{\rm train} = \frac{\Gamma_{\rm train}}{\Gamma_{\rm train}+\Gamma_{\rm sync}}T,
    \label{eq:training_rounds}
\end{equation}
where $\Gamma_{\rm train}$ and $\Gamma_{\rm sync}$ are the number of consecutive \tround and \sround rounds, respectively.
We define the training probability of a node $i$ as:
\begin{equation}
    p_i = \min \left (\frac{\tau_i}{T_{\rm train}},1\right),
    \label{eq:prob}
\end{equation}
where $\tau_i$ is the computational budget of node $i$.
Note that if the budget of a node $i$ is bigger than $T_{\rm train}$, i.e., $\tau_i \geq T_{\rm train}$, it is going to compute the updated model on each training round since $p_i = 1$, equivalent to \sys in the unconstrained settings.
By probabilistically skipping training, nodes in \sys spread their training budget across $T_{\rm train}$ rounds.
In summary, nodes in \sys-\constrained independently replace more \tround rounds with \sround rounds, in addition to the coordinated \sround rounds of \sys.

\Cref{fig:exp_comparison_fl} illustrates how \sys operates.
While \dpsgd trains at every round (\Cref{fig:exp_comparison_fl}a), \sys interleaves training rounds with coordinated syncing rounds (\Cref{fig:exp_comparison_fl}b).
Note that rounds labeled \emph{train} consist of train, share, and aggregate steps, whereas the ones labeled \emph{sync} consist of share and aggregate steps.
\sys-\constrained, in turn, leaves to each node the choice of whether to train or not during \emph{train} rounds, according to their energy budget.

\subsection{The \sys algorithm}
\Cref{alg:edpsgd} presents the pseudocode for \sys and \sys-\constrained executed in parallel by each node.
As mentioned earlier, \sys is a special case of \sys-\constrained, where $p_i = 1, \forall i \in V$.

\begin{algorithm2e}[h]
\caption{\texttt{\sys}, Node $i$}\label{alg:edpsgd}
\DontPrintSemicolon
Initialize $p_i$ with~\Cref{eq:prob}\;\label{line_edpsgd:start_init}
$\tau_i^1 \gets \tau_i$\;\label{line_edpsgd:end_init}
\For{$t = 1, \dots, T$}{
    $\xsym_i^{t-\frac{1}{2}} \gets \xsym_i^{t-1}$\;
    
    \If{$t \bmod (\Gamma_{\rm train} + \Gamma_{\rm sync}) < \Gamma_{\rm train}$ and $\tau_i^t > 0$ }{\label{line_edpsgd:train_condition}
        $r \gets$ random number between $0$ and $1$\; \label{line_edpsgd:prob}
        \If{$r \leq p_i$}{ \label{line_edpsgd:prob_condition}
            \For{$e = 1, \dots, E$}{\label{line_edpsgd:training_start}
                $\xi_i \gets$ mini-batch of samples from $D_i$\;
                $\xsym_i^{t-\frac{1}{2}} \gets \xsym_i^{t-\frac{1}{2}} - \eta \nabla f_i(\xsym_i^{t-\frac{1}{2}}, \xi_i)$\;
            }
            $\tau_i^{t+1} \gets \tau_i^t - 1$\
            \label{line_edpsgd:update_energy budget}\;
        }
    }
    Send $\xsym_i^{t-\frac{1}{2}}$ to the neighbors\; \label{line_edpsgd:share}
    $\xsym_i^t \gets \sum_{j\in V} W_{j,i} \xsym_j^{t-\frac{1}{2}}$\;\label{line_edpsgd:aggregate}
}
\KwRet $\xsym_i^T$\;\label{line_edpsgd:output}
\end{algorithm2e}
At the beginning of the execution, each node computes its training probability ($p_i$) following the~\Cref{eq:prob} and initializes the number of remaining training rounds based on its energy budget~(Lines~\ref{line_edpsgd:start_init}-\ref{line_edpsgd:end_init}).
The node then proceeds through a series of $T$ rounds.
In each round $t$, the node first decides if this is a coordinated \tround round or a coordinated \sround round~(Line~\ref{line_edpsgd:train_condition}).
Only if it is a coordinated \tround round, the node probabilistically decides if it wants to participate in training this round based on the energy budget~(Lines~\ref{line_edpsgd:prob}-\ref{line_edpsgd:prob_condition}).
If the node decides to participate in training, it performs the model update~(Lines~\ref{line_edpsgd:training_start}-\ref{line_edpsgd:update_energy budget}).
The node, in all cases, then moves on to sending the model to the neighbors and the aggregation step~(Lines~\ref{line_edpsgd:share}-\ref{line_edpsgd:aggregate}).
After executing $T$ rounds, the node outputs the final model~(Line~\ref{line_edpsgd:output}).
Note that if $p_i = 1$, nodes will always perform the training steps (model update) in each of the coordinated \tround rounds, and hence execute the unconstrained version of \sys.

\section{Evaluation}
\label{sec:eval}
We next describe the \sys implementation, experimental setup, the \sys hyperparameter optimization, and evaluation results.

\subsection{Implementation}
We implement \sys on top of the DecentralizePy framework~\cite{dhasade2023decentralized} in $700$ lines of Python 3.8.16 relying on PyTorch v.2.0.1~\cite{paszke2019pytorch} to train and implement \ML models\footnote{\url{https://github.com/sacs-epfl/SkipTrain.git}}. Each node is executed on a separate process, responsible for managing its dataset and executing tasks independently of other nodes. 

\begin{figure*}[ht!]
	\centering
	\includegraphics{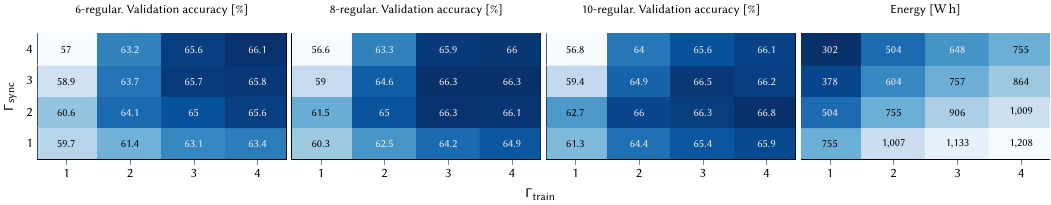}
	\caption{The average validation accuracy and energy consumption grid of \sys, for the \cifar dataset. Darker shades indicate better values.
    }
	\label{fig:heatmap_validation}
\end{figure*}

\subsection{Experimental setup}
\label{subsec:experimental_setup}

\paragraph{Cluster and network} We deploy our experiments on 8 machines with 2 Intel(R) Xeon(R) CPU E5-2630 v3 @ 2.40GHz of 8 cores and hyperthreading enabled. All the machines run Ubuntu 20.04.4 LTS with 5.4.0-99-generic kernel.
In our experiments, we emulate 256 nodes connected on $d$-regular topologies, with $d \in \{6,8,10\}$.

\paragraph{Datasets and hyperparameters} \sys is evaluated on two well-known image classification datasets: \cifar~\cite{krizhevsky2009learning} and \femnist~\cite{leaf}.
For the first dataset, we consider a $2$-shard \niid data distribution where images are sorted by labels and partitioned into $2n$ shards. Each node receives two shards, limiting the number of distinct labels per client~\cite{mcmahan2017communication}. 
\femnist, in turn, has a natural correspondence between dataset and nodes, \ie, all the handwritten characters produced by a user %
are clustered together. We pick the top-$256$ clients with the highest number of samples, and we map each to a node in our simulations.

We use \ac{CNN} architectures adapted from previous work~\cite{mcmahan2017communication, leaf, dhasade2023decentralized}.
These models are trained with \sgd and the Cross-Entropy loss function.
We tuned the learning rate ($\eta$) of each model with \dpsgd on a validation set obtained by extracting $50\%$ of the samples from the test set.
Hence, the validation and test sets are disjoint, containing $5000$ samples for \cifar and $20416$ samples for \femnist. 
The optimized hyperparameters for each dataset are reported in~\Cref{tab:sim_param}. 

\begin{table}[htb]
	\caption{Simulation hyperparameters for \cifar and \femnist datasets.}
	\label{tab:sim_param}
	\centering 
	\resizebox{\columnwidth}{!}{  
		\begin{tabular}{cccc}
			\toprule
			\textsc{Hyperparameter} & \textsc{Description} & \textsc{\cifar} & \textsc{\femnist} \\
			\midrule
			$\eta$ & Learning rate & $0.1$ & $0.1$\\
			$|\xi|$ & Batch size & $32$ & $16$\\
			$E$ & Local steps & $20$ & $7$\\
			$|\xsym|$ & Model size & $89834$ & ${1690046}$\\
			$T$ & Total number of rounds & $1000$ & $3000$ \\
			\bottomrule
	\end{tabular}}
\end{table}

\paragraph{Metrics} We evaluate the Top-1 accuracy on the validation and test sets, computed every $\Gamma_{\rm train} + \Gamma_{\rm sync}$ rounds.
We use the validation set to optimize our hyperparameters, including $\Gamma_{\rm train}$ and $\Gamma_{\rm sync}$ which are introduced by \sys.
We use the test set to determine model accuracies during all other experiments.

\begin{table*}[htb]
	\centering
	\caption{Energy traces for \cifar and \femnist datasets.}
	\label{tab:energy_traces}
	\begin{tabular}{c|cc|cc}
		\toprule
		\textsc{Device Name}                           & \multicolumn{2}{c}{ \textsc{Average Energy} [\SI{}{\milli\watt\hour}]} & \multicolumn{2}{c}{\textsc{Training rounds}} \\ 
		& \cifar & \femnist                                & \cifar & \femnist                                    \\ \midrule
		Xiaomi 12 Pro                         & \num[round-mode=figures]{6.532875} & \num[round-mode=figures]{21.508007} & \num{272} & \num{413}                   \\
		Samsung Galaxy S22 Ultra              & \num[round-mode=figures]{5.986076} & \num[round-mode=figures]{19.707797} & \num{324} & \num{492}                   \\
		OnePlus Nord 2 5G                     & \num[round-mode=figures]{2.556283} & \num[round-mode=figures]{8.415981} & \num{681} & \num{1034}                  \\
		Xiaomi Poco X3                        & \num[round-mode=figures]{8.51912}  & \num[round-mode=figures]{28.047269} & \num{272} & \num{413}                   \\ \bottomrule
	\end{tabular}
\end{table*}

\paragraph{Energy Traces}
Our analysis considers the total amount of energy consumed during the training process with each node $i$ in the network representing a smartphone. Our focus on smartphones stems from:
\begin{enumerate*}[label=\emph{(\roman*)}]
	\item their widespread adoption, which results in a vast amount of (often sensitive) data that can be leveraged for model training;
	\item suitability of efficient energy management for enhancing the end-user experience; and
	\item prior studies in \ac{FL}~\cite{bonawitz2019towards} have demonstrated the potential of large-scale \ac{ML} model training on smartphones.
\end{enumerate*}
Given these considerations, smartphones are an ideal setting to assess the effectiveness of energy-constrained \ac{DL} approaches.
In our evaluation, we consider networks consisting of four different smartphones: the Xiaomi 12 Pro, Samsung Galaxy S22 Ultra, OnePlus Nord 2 5G, and the Xiaomi Poco X3, and we derive the energy consumption by the training process for each device type.
Following \Cref{eq:energy_training_round}, we need information about the duration of the training process $\Delta_i^t$ and the power consumption $P_{\rm{hw}, i}^t$ for these smartphones.
We first derive the power consumption of each device type from the Burnout benchmark~\cite{ignatov:2022:burnout}. %
Then, we obtain the inference time of one data sample for \textsc{MobileNet-v2} from the AI benchmark~\cite{ignatov2018ai} for these device types.
We scale this inference time with the number of parameters in the model, local steps, and the batch size to get the total inference time.
Finally, we compute the training time following the methodology of the FedScale~\cite{lai2022fedscale}, \ie, scale the inference time with the batch size and a multiplier of $3 \times$.
We present in \Cref{tab:energy_traces} the energy spent by each device in one training round %
for each dataset.
Compared to \cifar, training on \femnist is more energy-demanding due to the larger model size. 

To further evaluate the performance of our approach in an energy-constrained setting, we obtain the maximum number of training rounds that can be executed by each device ($\tau_i$) as the number of rounds to exhaust a certain percentage of the battery capacity. 
We set this value to $10\%$ and $50\%$ for \cifar and \femnist, respectively.
In our simulations, we distribute the 256 nodes evenly among the four types of devices outlined in \Cref{tab:energy_traces}.

\subsection{Optimizing $\Gamma_{\rm train}$ and $\Gamma_{\rm sync}$}
\label{subsec:hyperparameters_optimization}

\sys introduces two new hyperparameters: $\Gamma_{\rm train}$ and $\Gamma_{\rm sync}$, which are the number of consecutive \tround and \sround rounds, respectively.
Choosing the right balance of \tround and \sround rounds is important for understanding and optimizing the performance of \sys.
We conduct a grid search using the \cifar dataset to tune these values for each network topology.
We choose the parameter combination that yields the highest average validation accuracy across all the nodes, with ties resolved in favor of the option with the lowest energy consumption.
Intuitively, we expect that as the degree of the considered network topology increases, the optimal value of $\Gamma_{\rm sync}$ will decrease.
This expectation arises from the understanding that fewer synchronization rounds are needed to converge to the average models across all nodes in more densely connected topologies.

\Cref{fig:heatmap_validation} shows the average validation accuracy and the total energy consumption (right-most heatmap) for combinations of $\Gamma_{\rm train}$ and $\Gamma_{\rm sync}$, over $1000$ rounds.
We note that the energy consumed during training exclusively depends on $T_{\rm train}$ and is independent of the network topology.
For the $6$-regular topology, the best validation accuracy is obtained when both $\Gamma_{\rm train}$ and $\Gamma_{\rm sync}$ are set to $4$, resulting in an average validation accuracy of $66.1\%$ and a total energy consumption of \SI{755}{\watt\hour}. 
For the $8$-regular topology, the best values for $\Gamma_{\rm train}$ and $\Gamma_{\rm sync}$ is $3$ with an average validation accuracy of $66.3\%$.
Interestingly, the same accuracy can also be reached with $\Gamma_{\rm train}=4$ and $\Gamma_{\rm sync}=3$, but this choice would require $12.4\%$ more energy. 
Unlike previous cases, for the $10$-regular topology, the highest validation accuracy of $66.8\%$ is achieved with $\Gamma_{\rm train} = 4$ and $\Gamma_{\rm sync} = 2$, with a total energy consumption of \SI{1009}{\watt\hour}. 
The energy consumption is greater than other topologies due to the number of training rounds $T_{\rm train}$, computed following \Cref{eq:training_rounds}, which is $666$ on the $10$-regular graph and $500$ in the other topologies. 

These experimental results support our intuition that by increasing the degree of the topology, the optimal value for $\Gamma_{\rm sync}$ decreases. 
The heatmap shows the effect of varying the number of synchronization rounds. Reducing it below the optimal values hinders the convergence of each node's model towards the global average, while increasing it beyond the optimal value leads to wasted execution rounds that could otherwise be allocated to training.
From an energy perspective, if we fix $\Gamma_{\rm sync}$, reducing $\Gamma_{\rm train}$ also reduces the energy requirements of the algorithm.
Therefore, the configuration with $\Gamma_{\rm sync} = 4$ and $\Gamma_{\rm train}=1$ requires only \SI{302}{\watt\hour}.
At the same time, this adversely reduces model accuracy. %

\begin{figure}[b!]
	\centering
	\includegraphics{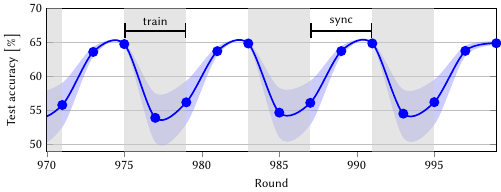}
	\caption{\sys average test accuracy with \cifar evaluated every $2$ rounds. The curve shadow indicates the standard deviation.
	}
	\label{fig:spikes}
\end{figure}

\subsection{The train/sync trade-off}
\label{subsec:trainSyncTradeoff}

In \Cref{fig:spikes}, we show the test accuracy results for \sys using a $6$-regular topology with \cifar. 
In this experiment, we evaluate the model every 2-rounds to understand the influence of training and synchronization rounds on the test accuracy. 
As expected, \sys shows a fluctuating behavior, \ie, the accuracy rises during synchronization rounds and drops during training rounds, while the standard deviation follows the opposite trend. 
This occurs because, during training, the model is biased towards the local dataset, which only contains samples for a subset of the available labels. In contrast, the test set follows an \iid distribution.
Therefore, the models obtained at the end of $\Gamma_{\rm train}$ training rounds are not optimized for the test set and significantly differ from the models of the neighbors, \ie, the average accuracy decreases while the standard deviation increases. During synchronization rounds, each node's model acquires knowledge from its neighbors enhancing its performance on labels not encountered during training, thereby improving test accuracy and reducing the standard deviation.

\begin{figure*}[t!]
	\centering
	\includegraphics{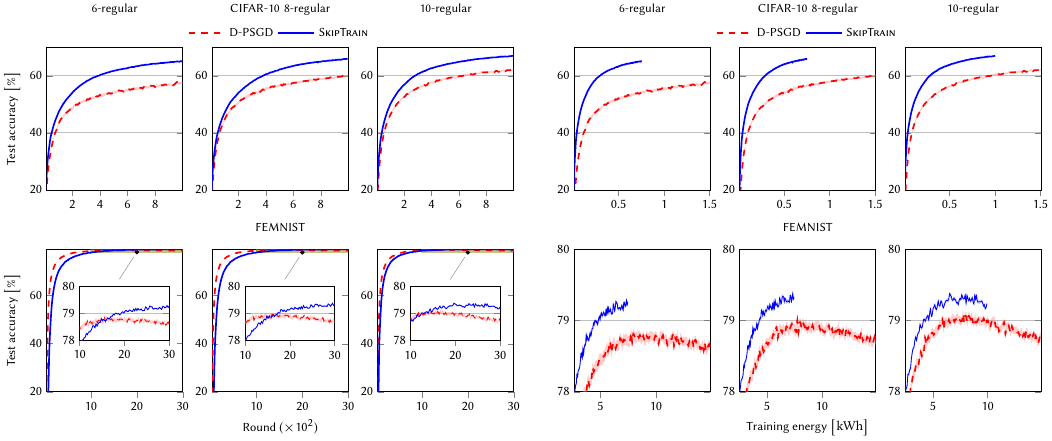}
	\caption{Comparing \sys to \dpsgd in terms of test accuracy and energy. %
	}
	\label{fig:dpsgd_comparison_cifar}
\end{figure*}

\subsection{\sys: performance}
\label{subsec:performance_syncronization_rounds}
In this section, we compare the accuracy and the energy consumption of \sys on a fixed number of total rounds $T$ on the \cifar and \femnist datasets.
\Cref{fig:dpsgd_comparison_cifar} shows the average test accuracy vs. rounds, and test accuracy vs. consumed training energy using the optimized combination of $\Gamma_{\rm sync}$ and $\Gamma_{\rm train}$. 
\sys consistently outperforms \dpsgd on \cifar by reaching on average $6\%$ higher accuracy across all topologies.
On the \femnist dataset, \sys reaches similar test accuracy values to \dpsgd while significantly reducing the energy consumption.
From the numerical results reported in \Cref{tab:edpsgd_no_probs_vs_dpsgd}, we observe that \sys consumes up to $2\times$ less energy on both datasets, since \sys performs half of the training rounds.
Finally, we observe little variation across different topologies.

\begin{table*}[]
\centering
\caption{Training energy consumption and average test accuracy for \sys and \dpsgd.}
\label{tab:edpsgd_no_probs_vs_dpsgd}
\begin{tabular}{cc|ccc|ccc}
\toprule
\textsc{Algorithm} & \textsc{Dataset}                   & \multicolumn{3}{c}{\textsc{Training Energy} [\SI{}{\watt\hour}]} & \multicolumn{3}{c}{\textsc{Average Test Accuracy [\%]}} \\
          &                           & 6-regular  & 8-regular & 10-regular      & 6-regular    & 8-regular   & 10-regular   \\\midrule
\sys      & \cifar                    & $755.02$   & $756.53$  &  $1008.71$      & $65.09$      & $65.93$     & $66.96$\\
\dpsgd    & \cifar                    & $1510.04$  & $1510.04$  & $1510.04$      &   $57.55$    & $60.08$     & $62.20$\\
\midrule
\sys      & \femnist                  & $7457.19$   & $7457.19$   & $9942.92$        & $79.26$   & $79.32$    & $79.24$       \\
\dpsgd    & \femnist                  & $14914.38$   & $14914.38$ & $14914.38$       & $78.6$        & $78.69$      &  $78.73$\\       
\bottomrule
\end{tabular}
\end{table*}
\begin{figure}[t!]
	\centering
	\includegraphics{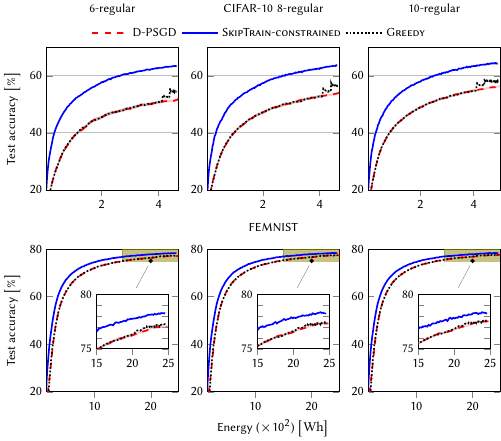}
	\caption{Comparison between \sys-\constrained, \greedy, and \dpsgd on the energy-constrained setting.
 }
	\label{fig:energy_accuracy_CIFAR}
\end{figure}
\subsection{\sys-\constrained: performance}
\label{subsec:performance_participation_probs}
In this section, we evaluate the performance of \sys-\constrained and the impact of training probabilities in %
the energy-constrained setting.
We continue to use the tuned values of the $\Gamma_{\rm train}$ and $\Gamma_{\rm sync}$ from \Cref{subsec:hyperparameters_optimization} and compare the test accuracy of \sys-\constrained with the two baselines having the same energy constraints: 
\begin{enumerate*}[label=\emph{(\roman*)}]
\item \dpsgd, and 
\item the \greedy algorithm as described in \Cref{subsec:partial_client_participation}, where each client executes training rounds until they exhaust their energy budget, and then only perform \sround rounds. %
\end{enumerate*}
Since \dpsgd is not inherently energy-aware, we do not limit it to individual energy constraints, \ie all devices in \dpsgd perform the training steps in all rounds.
On the other hand, for \greedy and \sys-\constrained, each node is associated with a smartphone (\Cref{tab:energy_traces}), and can only train while the training energy is not exhausted.

In \Cref{fig:energy_accuracy_CIFAR}, we present the test accuracy of each algorithm against the training energy consumed.
On the \cifar dataset, \sys-\constrained outperforms both \dpsgd and \greedy by reaching up to $12\%$ and $9\%$ higher accuracies, respectively.
On the \femnist dataset, the performance gap is smaller, but the trend remains the same.
We illustrate these results in \Cref{tab:edpsgd_probs_vs_dpsgd}.

It is important to note that the performance of \greedy against \dpsgd validates the benefits of \sround rounds.
The nodes in \greedy continue to share and aggregate when the training budget is exhausted and attain better accuracies than \dpsgd.

\subsection{\cifar and \femnist accuracies}
\begin{figure}
	\centering
	\includegraphics{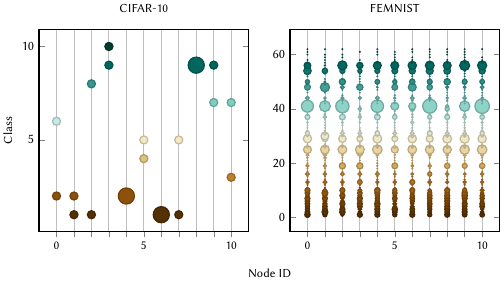}
	\caption{Class distributions for the first 10 nodes on \cifar and \femnist datasets. 
    Dot size corresponds to the number of samples of a given class in a given node.
 }
	\label{fig:bubles_datasets}
\end{figure}

As described in \Cref{subsec:performance_syncronization_rounds,subsec:performance_participation_probs}, it is evident that \sys is more energy-efficient than both \dpsgd and \greedy, while improving model performance.
However, on \cifar, the accuracy gap between \sys and \dpsgd is much more pronounced compared to \femnist. 
This discrepancy can be attributed to the distinct distributions of the two datasets.
As explained in \Cref{subsec:experimental_setup}, for \cifar, we adopted a $2$-shard data distribution, resulting in most clients having only $2$ out of $10$ possible classes within their datasets.
This results in a highly heterogeneous data distribution, hence producing less performant models~\cite{hsieh2020non}.
In contrast, the \femnist dataset is partitioned among the nodes such that samples written by the same person are clustered together~\cite{leaf}.
This results in a more homogeneous class distribution (different digits).
\Cref{fig:bubles_datasets} illustrates a comparison of the class distribution between \cifar and \femnist datasets for $10$ out of $256$ nodes, highlighting that \cifar presents a more challenging \niid class distribution. 
This difference justifies the smaller performance gap in terms of accuracy between \sys and \dpsgd in \femnist, when compared to \cifar.
This observation also holds true in an energy-constrained setting.

\begin{table*}[]
\centering
\caption{Training energy and average test accuracy for \sys-\constrained, \greedy, and \dpsgd in the energy-constrained setting.
}
\label{tab:edpsgd_probs_vs_dpsgd}
\begin{tabular}{cc|ccc|ccc}
\toprule
\textsc{Algorithm}              & \textsc{Dataset}                  & \multicolumn{3}{c}{\textsc{Energy budget} [\SI{}{\watt\hour}]} & \multicolumn{3}{c}{\textsc{Average Test Accuracy [\%]}} \\
                  &              & 6-regular  & 8-regular & 10-regular      & 6-regular    & 8-regular   & 10-regular   \\\midrule
\sys-\constrained & \cifar       & $462.7$   & $463.1$  & $490.55$          & $63.50$       & $63.52$     & $64.33$        \\
\greedy           & \cifar       & $463.37$  & $463.7$  & $491.18$          & $54.39$      & $56.57$     & $57.86$      \\
\dpsgd            & \cifar       & $468.11$   & $468.11$  & $498.31$        & $51.57$      & $53.98$     & $56.36$        \\
\midrule
\sys-\constrained & \femnist                  & $2455.43$  & $2454.97$ & $2454.29$       & $78.27$      & $78.26$     & $78.23$ \\
\greedy              & \femnist                  & $2460.41$   & $2460.41$  &  $1460.41$      &  $77.25$     & $77.45$    & $77.60$     \\
\dpsgd                  & \femnist                  & $2485.73$  & $2485.73$ & $2485.73$       & $77.05$      & $77.34$    & $77.54$       \\
\bottomrule
\end{tabular}

\end{table*}

\section{Discussion}
\label{sec:discussion}
We discuss the potential bias that might be introduced by our energy-aware approach, the implications of possible inaccuracies in the energy traces, along with  considerations of the practicality of \sys.

\subsection{Bias towards high-energy-capacity devices}
In the context of energy-efficient decentralized learning, it is crucial to recognize that a primary focus on energy efficiency can inadvertently bias the system towards high-energy-capacity devices. 
Low-energy-capacity devices in \sys will essentially skip more training rounds, leading to a situation where they contribute less to the collaborative learning process. 
On the contrary, high-energy-capacity devices will participate in more training rounds, thereby creating a bias towards their own local model.
This could exacerbate the performance gap between nodes where the converged model achieves disproportionately higher performance on the data of high-energy devices.
Therefore, while energy efficiency is a critical aspect of decentralized learning, it must be balanced with considerations of fairness and performance equity among participating devices. 
We leave the exploration of these ideas to future work.

\subsection{The accuracy of energy traces}
During our evaluation, we assumed that each node in the network represents a smartphone device and we created traces with training energy consumption to evaluate the effectiveness of \sys (see~\Cref{subsec:experimental_setup}).
However, as is the case with any trace-driven evaluation, our traces are an approximation of real-world behavior that might impact the accuracy and generalizability of the findings.
Primarily, the linear scaling of inference time based on model parameters, batch size, and local steps, although logical, might not precisely reflect the actual computational load and energy consumption on real smartphone devices.
Moreover, the approximation of training time as thrice the inference time, borrowed from the FedScale benchmark~\cite{lai2022fedscale}, is a simplification that may not hold across different smartphone models.
Our traces are also influenced by the accuracy of the Burnout benchmark~\cite{ignatov:2022:burnout}.
Despite the approximations introduced by our traces, we expect that \sys maintains its energy efficiency and superior model accuracy, even in deployed settings.

\subsection{Practical implications of \sys}

\dpsgd is a well-established \ac{DL} algorithm, but it presents practical limitations owing to the static topology and synchronous nature when deployed in real-world and large-scale networks.
While the concept of skipping \tround rounds in favor of \sround rounds is generally applicable in decentralized learning, we take \dpsgd as a foundation for \sys.
We hence inherit both the benefits and limitations of \dpsgd, including its synchronous operation.
\sys uses coordinated synchronization rounds to reduce the energy consumed, which can be effectively implemented as an extension of \dpsgd.
Such coordination can be however more challenging to implement at scale.
In contrast, asynchronous algorithms~\cite{lian2018asynchronous} offer a more practical approach by relaxing the need for strict synchronization. 
We leave the exploration and development of an asynchronous extension of \sys for future research.

\section{Related work}
\label{sec:related}

The energy footprint of \ac{ML} algorithms is an increasing concern, with many works trying to estimate and reduce this footprint~\cite{garcia2019estimation,kumar2022optimization,wu2022sustainable,savazzi2021framework,strubell2019energy}.
We first describe energy-aware techniques that reduce the power consumption of training in \ac{FL} and \ac{DL}, two popular approaches for distributed \ac{ML}.
We then discuss efforts that, similarly to \sys, leverage additional communication rounds and partial node participation.

\textbf{Energy-Efficient \ac{FL}.} Various techniques in the domain of \ac{FL} have been proposed to reduce the energy consumption of end devices, \eg, by intelligently selecting workers or by optimizing the local training process~\cite{qiu2023first}.
Arouj \etal~\cite{arouj2022towards} have introduced an energy-aware \ac{FL} mechanism that selects clients with higher battery levels,  maximizing system efficiency.
Similar to \sys, they evaluate their approach using energy traces of real devices.
AutoFL is a reinforcement learning-based method that optimizes \ac{FL} by strategically selecting participating devices and setting per-device execution targets each aggregation round, enhancing model convergence time and energy efficiency~\cite{kim2021autofl}.
Most of these algorithms modify the client selection procedure and are therefore unsuitable in the \ac{DL} setting assumed by \sys due to the lack of centralized control.
FLeet is an online learning system that reduces the energy impact of the learning task on mobile devices by profiling the device energy consumption and adapting the batch size accordingly~\cite{damaskinos2022fleet}.

\textbf{Energy-Efficient \ac{DL}.}
Much effort in the domain of \ac{DL} has been put into reducing the communication volume, \eg through sparsification~\cite{dhasade2023get}.
While this might reduce power consumption and overall training time, only a few studies aim to directly reduce and measure the energy impact of \ac{DL} algorithms.
Aketi et al.~\cite{aketi2021sparse} propose Sparse-Push, an energy-efficient \ac{DL} algorithm that reduces training and communication overhead, therefore reducing energy usage.
Several works formally model the overall power consumed during the learning process and propose techniques to reduce energy impact, \eg, through decentralized aggregation with overlapping clusters of peers~\cite{al2023decentralized} or by strategically selecting which peers will transmit their updated model~\cite{liu2023communication}.
Tang et al.~\cite{tang2021battery} study how to optimize federated edge learning in UAV-enabled \ac{IoT} for cellular networks in which devices have, similar to our setting, battery constraints.
We are the first to empirically verify the performance of \ac{DL} in energy-constrained settings, to the best of our knowledge.

\textbf{Multiple Gossip Steps (MGS).}
Several works have studied the advantages of performing multiple gossip steps between subsequent local model updates~\cite{shi2022improving,li2022destress}.
Scaman \etal~\cite{scaman2017optimal} analyze a distributed optimization method where nodes use MGS to reduce the variance between their models.
Their work, however, evaluates the effectiveness of MGS on simple linear problems.
The NEAR-DGD+ algorithm~\cite{berahas2018balancing} is an optimization method that increases the number of communication steps per round as the network makes progress.
Network-DANE is a decentralized algorithm that adopts multiple rounds of mixing within each iteration, which helps accelerate convergence when the network exhibits a high degree of locality~\cite{li2020communication}.
Hashemi \etal~\cite{hashemi2021benefits} theoretically analyze \dpsgd under a fixed communication budget and propose to use compression and multiple gossip steps between subsequent gradient iterations to increase the efficacy of learning.
The Muffliato \ac{DL} algorithm~\cite{cyffers2022muffliato} alternates between multiple rounds of gossip and local training.
However, the main objective of Muffliato is to enhance privacy in \ac{DL} through local noise injection, whereas \sys primarily introduces sharing-only rounds to reduce power consumption.
We remark that most existing works adopting MGS are primarily theoretical contributions, and unlike our work does not not integrate an energy model or evaluate the effectiveness of their approach with deep neural networks.

\textbf{Partial Client Participation.}
\sys uses an energy-aware, probabilistic method to determine if a client should participate in a particular round or not (see~\Cref{subsec:partial_client_participation}).
Client selection in \ac{FL} is a well-studied problem and there have been various proposals to strategically select clients to reduce time-to-accuracy or increase model accuracy~\cite{nishio2019client,lai2021oort,cho2022towards,fu2023client}.
However, these methods cannot be applied to \sys as they rely on statistics collected by the central server.
Several papers consider partial client participating in \ac{DL} settings.
Guerra \etal~\cite{guerra2023fully} propose a decentralized client selection policy for \ac{FL}, based on non-stationary multi-armed bandits where clients make a local decision each round on whether to participate or not in the training.
Liu et al., prove the convergence and effectiveness of \dpsgd when sub-sampling a small set of nodes each round that train the model~\cite{liu2022decentralized}.

\section{Conclusion}
\label{sec:conclusion}
In this work, we introduced \sys, a novel \ac{DL} algorithm devised to reduce energy consumption in decentralized learning environments by selectively substituting training rounds with synchronization rounds, saving energy while enhancing model performance.
Through comprehensive empirical evaluations across 256 nodes, our findings demonstrated that \sys effectively reduces energy consumption by 50\% while elevating model accuracy by up to 7 percentage points compared to the conventional D-PSGD algorithm. Additionally, we extended our approach to \sys-\constrained for settings with individual energy budgets, such as \ac{IoT} or \ac{UAV}, where it showcased an improvement in model accuracy by up to 12 percentage points against D-PSGD. By integrating realistic energy traces in our evaluations, we provided a pragmatic measure of the energy savings, underscoring \sys as a promising approach for energy-efficient decentralized learning.%

\bibliographystyle{ACM-Reference-Format}
\bibliography{main_beautified}

\end{document}